\documentclass[twoside,leqno,twocolumn]{article}

\usepackage[letterpaper]{geometry}
\usepackage[table,x11names]{xcolor}
\usepackage{ltexpprt}
\usepackage{hyperref}
\usepackage{caption}
\usepackage{graphicx}
\usepackage{makecell}
\usepackage{changepage}
\usepackage{booktabs}
\usepackage{multirow}
\usepackage{balance}

\begin{document}

\title{\Large Betting the system:\\ Using lineups to predict football scores
}
\author{George Peters\thanks{Computer Science Department, University of Exeter, UK. \{gp421, D.Pacheco\}@exeter.ac.uk}
\and Diogo Pacheco\footnotemark[1]}

\date{}

\maketitle


\fancyfoot[R]{\scriptsize{Copyright \textcopyright\ 2023 by SIAM\\
Unauthorized reproduction of this article is prohibited}}





\begin{abstract} \small\baselineskip=9pt 

This paper aims to reduce randomness in football by analysing the role of lineups in final scores using machine learning prediction models we have developed. 
Football clubs invest millions of dollars on lineups and knowing how individual statistics translate to better outcomes can optimise investments.
Moreover, sports betting is growing exponentially and being able to predict the future is profitable and desirable. 
We use machine learning models and historical player data from English Premier League (2020-2022) to predict scores and to understand how individual performance can improve the outcome of a match.
We compared different prediction techniques to maximise the possibility of finding useful models.
We created heuristic and machine learning models predicting football scores to compare different techniques.
We used different sets of features and shown goalkeepers stats are more important than attackers stats to predict goals scored.
We applied a broad evaluation process to assess the efficacy of the models in real world applications.
We managed to predict correctly all relegated teams after forecast 100 consecutive matches.
We show that Support Vector Regression outperformed other techniques predicting final scores and that lineups do not improve predictions. 
Finally, our model was profitable (42\% return) when emulating a betting system using real world odds data.

\end{abstract}

\section{Background and Motivation}
Football is one of the most popular sports in world, if not the most~\cite{giulianotti1999football}. 
It is played and watched by millions worldwide.
The sport is also a popular domain for Data Science research.
The scientific community has explored performance indicators, such as how to make better assessments~\cite{thakkar2021assessment} or perform efficient team selection~\cite{rajesh2020data}.
Network models have been used to evaluate performance~\cite{cintia2015network}.
And social media data has been used to 
characterise supporters~\cite{pacheco2015football}, to understand rivalries~\cite{pacheco2016characterization} and to predict crime~\cite{pacheco2017using}.

Football analytics, on the other hand, is a small field when compared to other large sports, especially those based primarily in America like Baseball and Basketball. 
This is not caused by a lack of holistic investment in football.
The Premier League recently announced an extension of their TV rights deal with a number of broadcasting companies worth £5.1bn~\cite{PremierNews}. 
The football data analytics industry whilst small is growing, the recent success of Brentford in the Premier league has been largely attributed to their data driven ``Moneyball'' approach~\cite{SoccersBloomberg}. 
This seems to be indicative of a trend, during the recent takeover of Chelsea their new owner Todd Boehly expressed his disappointment at the data analysis structures in place and emphasised the importance of them for the club moving forward~\cite{ToddMore}. 

When one attempts to predict football results, the first decision is whether to handle the problem as a classification -- predicting the outcome, i.e., win, draw, or loss -- as a regression -- predicting the final score, e.g., 2:1 to the home team.
Academics have been applying statistical analysis to analyse the distribution of goals and use this distribution to predict outcomes. 
M. J Maher~\cite{Maher1982ModellingScores} was one of the first modelling the association football scores, in 1982.
He used an independent Poisson model to describe football scores. 
Contemporary work in this field is also rather extensive~\cite{arntzen2021predicting,maozad2022comparative,2018INDIVIDUALTechniques,PredictingXplore}.
Hucaljuk and Rakipović~\cite{PredictingXplore} use classification models to predict the outcome of Champions league matches. 
Their models were able to achieve accuracies of 68\% and 65\%. 
Horvat and Job~\cite{Horvat2020TheReview} evaluate the state of football prediction analysing the outcomes of 20+ models. 
They show classification models consistently attaining better results suggesting it to be an easier task.
Lindberg and Söderberg~\cite{lindberg2020comparison} compared a number of regression models with the aim of predicting an individual players performance during a match. 
They used long short-term memory neural networking and averaged MSE of 8.63 across the position groups. 
In 2018, Herbinet used in game stats, like expected goals (xG), to developed a classification model and a regression model~\cite{2018INDIVIDUALTechniques}. 
Their regression model was able to achieve an RMSE of 1.153 and an MAE of 0.861. 

This paper focus on football score prediction.
We implement regression models to predict home and away goals scored, independently.
We exploit machine learning regression techniques to find the best fit model, but we also implement naive heuristic models to have broader evaluation scenarios.
To the best of our knowledge, Herbinet's~\cite{2018INDIVIDUALTechniques} is the most similar work to ours in terms of task and features. 
The main difference is our paper investigates the importance of \emph{individual} player lineup statistics rather than global measurements for an entire team.



\section{Methods and Technical Solutions}

\subsection{Problem}

We propose to predict football final scores by creating two independent models -- home goals scored and away goals scored. 
The final score is the combination of these predictions resulting in the final scoreline. 
We test different approaches to understand which technique can generate the best results, but we want to incorporate lineup data.
First, to see if they can produce better predictions.
But more importantly, to unveil how individual player statistics affect a match outcome.

\subsection{Evaluation Framework}
\label{sec:evaluation}
We propose a comprehensive evaluation, assessing models fitness, real world scenarios, and simulating bets.
Fitness is measure by the commonly used Mean Absolute Error (MAE), Root Mean Squared Error (RMSE), and R Squared (R2)~\cite{Wu3Science}.

Three real world scenarios are evaluated: (i) overall standings goodness based on Kendall $\tau$ rank correlation; (ii) accuracy prediction of top four teams (UEFA Champions League group stage); and (iii) accuracy of bottom three teams (relegation).
Each score predictions in the test set is combined into match scoreline, and we compute the table standings assigning three points for a win, one point for a draw, and zero for a loss. 
This table is then compared to the actual premier league points earned by each team over the same period. 

We simulate consecutive bets, starting with a pot of £0. 
For each game in the test set, we bet £1 per each model predicted score. 
If the prediction is correct, the stake is multiplied by the odds\footnote{Obtained at \texttt{oddsportal.com}.} and added to the models pot.

\subsection{Datasets}
The data consists of 680 matches from the English Premier league from seasons 2020, 2021, and 2022.
Matches are sorted chronologically, and split in training and test (last 100 matches).
We combined two datasets to train and test our models, one providing football fixtures\footnote{See \texttt{fixturedownload.com}.} and the other containing individual players statistics\footnote{Scrapped from \texttt{FBRef.com}.}. 

The list of stats available is vast, and specific for player's position. For instance, goals against and saves for goalkeepers; and long passes completed and assists for outfield players. 
Stats can be further classified as defensive (e.g., tackles won) or offensive (e.g., shots on target).
Due to space constraints, we omit the the full list of stats here, but made it available online\footnote{\texttt{github.com/georgejpeters/Feature-list-and-legend}}.


\subsection{Models}
We propose six different approaches, three simple naive heuristics and three distinct combination of features.
The latter approaches are implemented using five machine learning techniques, thus resulting in a total of eighteen models.
They are detailed in the following subsections.
This configuration allows us to understand the effect of lineups and various machine learning techniques on the accuracy of predictions. 


\subsubsection{Heuristics}
These first three models do not implement any machine learning techniques and are therefore useful baselines in analysing the general effectiveness of ML models on predicting the result of football matches. 

\paragraph{Home Win}
The first baseline model simply predicts a 1:0 home win for every fixture, whilst this is a naive model it is motivated by the fact that 1:0 is the most common scoreline in the premier league at 16.1\% \cite{WhatOffers}.

\paragraph{Tradition}
The second baseline model predicts a 1:0 win for whichever team is higher in the table at the end of the training set. If Chelsea is $3^{rd}$ in the table and West Ham is $8^{th}$, the model predicts 1:0 Chelsea win. 
This model is very brittle and does not deal with teams like over-performing well, but it gives a general simulation of the average punter's approach to score prediction.

\paragraph{Recency}
The third baseline model creates the scoreline using each teams goals scored in the immediate previous match (repeating the past). 
For example, if in their $n^{th}$ match, Arsenal won 1-0 and Chelsea won 3-1, and the $(n+1)^{th}$ match is Arsenal \textit{vs.} Chelsea, the predicted score is 1-3 Chelsea to win. 
This approach investigates the intuition that the past repeats itself and that teams who have performed well in very recently will keep their perform. 

\subsubsection{Combining Features}
The next three approaches allows us to investigate the importance of different sets of features in predicting score.

\paragraph{Players}
The first approach disregard all statistics and uses only the player lineups.
The features used by the machine learning models are the encoded names of the all players present in the training set. 
Each player is a feature which is set to 1, -1 or 0 representing whether they played for the home team, away team or neither, respectively. 
This approach enables us to investigate the potential impact of individual based on their social capital rather than their objective productive indicators. For instance, does the presence of Cristiano Ronaldo impacts the results regardless of his form?

\paragraph{Lineup Stats}

Each model is trained using aggregated game data from the current season and the season previous.
First, the individual players listed on the lineup of a match have their stats are aggregated into the average of each position groups stats (goalkeepers, defenders, midfielders, attackers) up to that point in the season.
The features encompass 13 defenders stats, 14 midfielders stats, 13 attackers stats, and 12 stats from the opponent team (5 from goalkeeper and 7 from defenders).
In total, 52 features are used.
As a scoreline consists of two separate models (home and away goals scored) the home goals scored model has the offensive stats for the home team and the defensive stats for the away team and vice-versa. 
For example, a schema input into the home team goals scored model will have home team's midfielders goal creating actions, home team midfielders xG, away keeper's clean sheets, to name a few.

\paragraph{Team Stats}
The last approach is very similar to the previous one and also uses 52 features.
However, these features are calculated differently.
Team Stats models disregard lineups and average the statistics for the entire squad. 

\subsubsection{Machine Learning Techniques}
In order to instantiate \emph{Players}, \emph{Lineup Stats}, and \emph{Team Stats} approaches, we implement five ML techniques.

\paragraph{Linear regression (LN)}
is the simplest of the models.
It takes all the given data and tries to find the appropriate weightings for each feature and from that tries to map a line of best fit to the observed data.
It then uses this line of best fit to find the dependent variable when given a set of unseen data. 
In a home goals scored model the dependent variable would be how many goals the home team scores whilst the independent variables could be, home team midfielders key passes, home team attackers xG and away team goals conceded. 
Linear regression is useful for this paper's goals as somewhat of a baseline as it is by far the simplest model. 
Therefore, we can compare the more complex models to it to find out whether the added complexity is justifiable or merely a hindrance.

\paragraph{K-nearest neighbour (KNN)}
is a non-parametric model so it produces predictions using feature similarity, its predictions are based on which datapoint(s) query data most resembles in the training set \cite{K-NearestPython}. In regression problems a k number is input, this number dictates how many of the nearest points the algorithm should use for the final prediction, so if k=3 the three nearest points to the prediction datapoint will be selected. These nearest datapoints are then averaged to create the continuous value for the prediction. The distance from the prediction datapoint and the rest of the data is measured in a number of ways depending on the type of data in the model. KNN was selected because provides some model variety as it's the only non-parametric model. This is desirable as developing many similar models often leads to homogeneous results. In our project the model will be averaging k number of goal prediction values and returning the result.

\paragraph{Decision Tree Regression (DTR)}
are predictive models that use a set of binary rules to calculate a target variable \cite{DecisionDepth}. 
In this project the target variable is goals scored and the binary rules are related to the schema features input, for example home midfielders xG.A decision tree regression (DTR) model splits by asking questions of the data, i.e. are home attackers xG \textgreater~0.5, these questions are formed when training the model using the training data. To contextualise this process for this project 4 binary rules for a home goals scored model might be home attackers xG, home midfielders key passes, away defenders tackles won and home midfielders xA.

\paragraph{Random Forest Regression (RFR)}
is an ensemble machine learning technique, this means the model seeks to achieve better performance by combining the predictions of multiple models. 
In the case of RFR it uses multiple decision tree regression algorithms and combines them in a process called bagging \cite{RandomDepth}. In this project's case five bagged decision trees could predict a home goals scored of 0.8, 1.2, 1.5, 0.9 and 1.1, therefore, the overall prediction would be 1.1 (rounded to 1). Random forest regression is useful on its own to solve regression problems but also to investigate the impact of bagging on model accuracy.

\paragraph{Support Vector Regression (SVR)}
discovers the hyperplane with the most number of points in an n-dimensional space, where n is the number of features, this hyperplane will assist us in in predicting a discrete value (goals scored). 
In the case of this project, SVR finds the hyperplane that intersects the largest number of points in a 52 dimensional space. This hyperplane is then used to predict goals scored when presented with new data, ignoring any data points that fall outside of the decision boundaries.
SVR is used in to predict individual player performance~\cite{lindberg2020comparison} and to generate xG data~\cite{2018INDIVIDUALTechniques}.
It is a relatively complex algorithm so we can investigate whether the added complexity contributes to an increase in accuracy or if simpler models can achieve similar results.  

\section{Empirical Evaluation} 
The evaluation of the team stats and lineup stats models takes three forms, firstly the evaluation metrics RMSE, MAE and R2 are employed. These metrics are most useful for comparing the ML models against each other and hyper-parameter tuning. The next two evaluation  techniques are used to contextualise the models in the real world so we can better visualise their performance. Analysing each models performance in the betting market allows us to analyse whether the models outperform odds setters, who themselves use ML techniques to set such odds. Finally, building a points table shows us how each models perform in a slightly more forgiving arena (outcome prediction) and see if it can effectively predict teams performance over an extended period. 
    
\subsection{Model Fitness}


\begin{table}
    \centering
    \begin{tabular}{l| ccc} 
         \toprule
         Model & MAE & RMSE & R2 \\ 
         \midrule
         Home Win & 1.12 & 1.42 & -0.23 \\
         Tradition & 1.0 & 1.46 & -0.30 \\
         Recency & 1.41 & 1.83 & -1.06 \\
         Players LR & $<<1$ & $<<1$ & -1.97\\ 
         Players KNN & 1.05 & 1.37 & -0.14 \\ 
         Players DTR & 0.97 & 1.29 & -0.02 \\
         Players RFR & 0.97 & 1.32 & -0.07 \\
         Players SVR & 1.15 & 1.50 & -0.38\\
         Lineup Stats LR & 0.93 & 1.22 & 0.09\\ 
         Lineup Stats KNN & 0.89 & 1.26 & 0.03 \\ 
         Lineup Stats DTR &  0.95 & 1.34 & -0.09 \\
         Lineup Stats RFR & 0.92 & 1.22 & 0.10 \\
         Lineup Stats SVR & 0.89 & \textbf{1.16} & \textbf{0.17} \\
         Team Stats LR & 0.93 & 1.22 & 0.09\\ 
         Team Stats KNN & \textbf{0.86} & 1.23 & 0.07 \\  
         Team Stats DTR & 0.96 & 1.34 & -0.10 \\
         Team Stats RFR & 0.93 & 1.22 & 0.09 \\
         Team Stats SVR & 0.94 & 1.23 & 0.07 \\
         \bottomrule
    \end{tabular}
    \caption{\emph{Home} score prediction models results. Best results are highlighted.}
    \label{table:home_fitness}
\end{table}

\begin{table}
    \centering
    \begin{tabular}{l| ccc} 
         \toprule
         Model & MAE & RMSE & R2 \\ 
         \midrule
         Home Win & 0.94 & 1.30 & -0.06 \\
         Tradition & 1.08 & 1.52 & -0.43 \\
         Recency & 1.41 & 1.76 & -0.93 \\
         Players LR & $<<1$ & $<<1$ & -2.35\\ 
         Players KNN & 1.15 & 1.47 & -0.35 \\ 
         Players DTR & 0.94 & 1.33 & 0.09\\
         Players RFR & 0.97 & 1.34 & -0.11 \\
         Players SVR & 1.06 & 1.40 & -0.22\\
         Lineup Stats LR & 0.94 & 1.22 & 0.08\\ 
         Lineup Stats KNN & 0.94 & 1.33 & -0.11 \\ 
         Lineup Stats DTR & 0.98 & 1.36 & 0.16 \\
         Lineup Stats RFR & 0.95 & 1.24 & 0.05 \\
         Lineup Stats SVR & 0.90 & 1.17 & 0.16 \\
         Team Stats LR & 0.94 & 1.22 & 0.08\\ 
         Team Stats KNN & 0.91 & 1.34 & -0.11 \\  
         Team Stats DTR & 0.97 & 1.36 & -0.15 \\
         Team Stats RFR & 0.95 & 1.24 & 0.05 \\
         Team Stats SVR & \textbf{0.87} & \textbf{1.14} & \textbf{0.20} \\
         \bottomrule
    \end{tabular}
    \caption{\emph{Away} scores prediction results. Best results are highlighted.}
    \label{table:away_fitness}
\end{table}

Table~\ref{table:home_fitness} and~\ref{table:away_fitness} show the different evaluation metrics for the home and away prediction models, respectively. 
When comparing the machine learning techniques, Support vector regression seems to perform slightly better than the others for both home and away models.
When comparing away \emph{vs.} home models, the former tends to outperform their counterpart.
Whilst the deltas between evaluation metrics are generally not massive, it is worth to highlight the difference of Team Stats SVR R2 values.
Moreover, the results suggest that taking lineups into account is more impactful on the accuracy of models predictions at home than away. 
A possible explanation is that teams are more predictable at away grounds and are less influenced by lineups. 
In a similar vein it could suggest that lineups are more similar at away grounds due to managers needing to always play their best team, and therefore they have less impact on the models predictions. 
The improvement in the model in these scenarios could therefore be explained by access to a larger pool of data (as players that don't start aren't filtered out) causing more accurate predictions. However, as the delta is generally small between the performance metrics none of these theories can be definitively proven, but it would seem that taking lineups into account does have an impact on model's prediction accuracy.

\subsection{Feature Importance}
A further evaluation carried out on the lineup stats model was ranking feature importance, to do this we used chi-squared. Chi-squared measures the dependence between stochastic variables \cite{Sklearn.feature_selection.chi2Documentation}, using this test the function finds the features that are best at predicting correct results (in our case goals scored). Table~\ref{tab:best_features} shows the 5 most important features in the lineup stats model. 
Each feature represents an average up to that point in the season for the players in that position group.

\begin{table*}
    \centering
    \begin{tabular}{c m{1.2cm} m{4.3cm} r c m{1.2cm} m{4.3cm} r}
      \toprule
      \multirow{2}{*}{Rank} &
      \multicolumn{3}{c}{Home} &
      &
      \multicolumn{3}{c}{Away} \\ 
      \cmidrule{2-4} \cmidrule{6-8}
      & Feature & Description & Value &&
      Feature & Description & Value \\
	\midrule
	1 & away g\_CS & Clean sheets for the away teams goalkeeper. & 420.0  &
	& g\_CS & Clean sheets for the home team's goalkeeper. & 405.0 \\
	
	2 & away g\_GA & Goals against for the away teams goalkeeper. & 80.5 &
	& g\_GA & Goals against the home team's goalkeeper. & 96.5 \\
	
	3 & a\_GCA & Goal creating actions for the home team's attackers. & 76.1 & 
	& away a\_GCA & Goal creating actions for the away team's attackers. & 60.1\\
	
	4 & m\_GCA & Goal creating actions for the home team's midfielders. & 67.9 &
	& away m\_Gls & Goals scored by the away team's midfielders. & 55.1\\
	
	5 & m\_Gls & Goals scored by the home team's midfielders. & 64.1 &
	& g\_PSxG & Post shot expected goals for the home team's goalkeeper. & 53.6\\
	
	\bottomrule
    \end{tabular}
    \caption{Most important features for both \emph{Home} and \emph{Away} models.}
    \label{tab:best_features}
\end{table*}

These feature importance rankings provide valuable insights into the best predictors for a matches final score. To be clear these are not the features that will cause a team to score the most goals, rather these features are the best indicators for predicting the correct number of goals scored. Our intuition at the outset of this project was that average goals scored for each position group would dominate the feature importance, this turned out not to be the case. Instead goalkeeper stats take the top two importance spots for both the home and away model, with a keeper's clean sheets (not conceding a goal in a match) chi2 score being higher than the next four features combined. This seems to show that the quality of the opposition team's keeper is by far the most important factor when predicting how many goals a team will score. Now this is most likely for a few reasons, firstly keepers' importance are historically under emphasised by fans and pundits alike. World class managers like José Mourinho have made careers off organising impenetrable defenses, with the philosophy being if they can't score you can't lose. In the 2004/05 season his Chelsea team conceded only 15 goals, they went on to win the league with a record breaking number of points, losing only one game. It seems clear that the quality of a team's keeper has a large influence on how many goals the opposition will score. However, stats like a goalkeepers clean sheets are not determined solely by the keeper, the whole team's defensive efforts contribute. In Chelsea's 2004/05 season not only did they have the immense Petr Cech in goal, in front of him was a world class defensive spine of John Terry at centre back and Claude Makélélé in defensive midfield. A goalkeeper's clean sheets and goals against are a reflection of teams defensive abilities as a whole. Furthermore, it is the best most centralised stat to reflect a team's defensive capabilities. Other stats like defensive tackles won or shots blocked distribute a teams defensive abilities over many position groups and many features. Goalkeeper's clean sheets and goals against are uniquely positioned to reflect the team's defensive efforts due to the nature of how other stats are distributed. It must be noted that the away feature ranking had a third keeper attribute in the top five. This would suggest that a keeper's quality is even more important for predicting how many goals an away team will score.

The next interesting aspect of the feature rankings is the offensive stats that seem to be the most important for prediction accuracy. Attackers goal creating actions and midfielder goals were in the top five for both models, with midfielder goals only making it into the home model's top five. Attackers goals seems conspicuously absent, our intuition at the outset was that this stat would dominate feature importance. This is mostly due to conventional wisdom placing a large emphasis on having an attacker getting 20+ goals a season. The largest teams clearly share this belief as four out of five of the most expensive transfers ever are for such players \cite{TheUK}. Instead, emphasis is placed on goal creating actions, these are actions like passes, crosses and dribbles that lead to goals. These stats represent a teams creativity and shows that a it has a significant impact on a models goals scored prediction.

\subsection{Real World Scenarios}

\begin{table}
    \centering
    \begin{tabular}{l|cc} 
         \toprule
         Model & $\tau$\\
        \midrule
         Players & \textbf{0.232}\\
         Team Stats & 0.053\\
         Recency & 0.042\\ 
         Lineup Stats & -0.021\\ 
         Home Win & -0.105\\
         Tradition & -0.168\\  
          \bottomrule
    \end{tabular}
    \caption{Models rank correlations of the general standings table.}
    \label{table:tau}
\end{table}

The next form of evaluation carried out on the models involved converting the score predictions to outcome predictions as described in Section~\ref{sec:evaluation}.
Table \ref{table:tau} shows the rank correlation results. All the models performed poorly in trying to predict the entire table, however, this is somewhat to be expected as ranking all 20 teams perfectly is an extremely onerous task. 
If even one team over or under performs the rest of the teams rankings get shifted. 
Despite being the best model in this scenario, Players did not perform as well in model fitness.

The next evaluation involves finding the percentage of top four and bottom three correctly predicted (order disregarded). This is because the bottom three teams in the Premier league are relegated and the top four teams  get to play Champions league football. The difference between a team finishing 11th and 12th is somewhat inconsequential, but the difference between 4th and 5th can make or break a season. As SVR performed the best on the performance metrics it is the ML implementation we will use for all ML based models.

\begin{table}
    \centering
    \begin{tabular}{l|c c} 
         \toprule
         Model & Top-4(\%) & Relegation(\%) \\  
         \midrule
         Lineup Stats & \textbf{50} & \textbf{100} \\ 
         Team Stats & \textbf{50} & \textbf{100}  \\
         Players & \textbf{50} & 67\\
         Tradition & \textbf{50} & 33\\  
         Recency & \textbf{50} & 0\\  
         Home Win & 25 & 0\\
          \bottomrule
    \end{tabular}
    \caption{Models accuracy predicting the top-4 Champions League group and the bottom-3 relegated.}
    \label{table:t4}
\end{table}

Predicting the top-4 seems to be an easy task since even the heuristic models were able to predict 50\% correctly.
For the relegated teams, on the other hand, machine learning significantly improved the predictions.
Interestingly, predicting top and bottom is not direct correlated to predict the whole table.
The lineup model performed really well in the former, but not in the latter.

\subsection{Betting}

\begin{table}
    \centering
    \begin{tabular}{l| c} 
         \toprule
         Model & Betting Earnings (£) \\ 
         \midrule
         Team Stats KNN & \textbf{42.5}  \\  
         Lineup Stats DTR & 18.9\\
         Team Stats DTR &  18.8 \\
         Lineup Stats KNN & 14.6 \\ 
         Home Win & -7.0\\
         Tradition & -16.4\\
         Team Stats SVR & -16.8 \\
         Team Stats LR & -19.4 \\ 
         Team Stats RFR & -19.4 \\
         Lineup Stats LR & -19.4 \\ 
         Lineup Stats RFR & -19.4 \\
         Lineup Stats SVR & -30.9  \\
         Players KNN & -37.3 \\
         Players DTR &  -48.8 \\
         Players RFR & -50.1 \\
         Players SVR & -55.6 \\
         Recency & -62.5\\
         Players LR & -90.0\\ 
         \bottomrule
    \end{tabular}
    \caption{Models earnings after predicting and betting £1 in 100 consecutive matches.}
    \label{table:earnings}
\end{table}

The last evaluation carried out on our models centered on analysing their performance in the betting market.
Table~\ref{table:earnings} shows KNN and decision tree regression managed to make profits in both the Team stats model and the lineup and stats model. 
With the largest profits being made by the team stats KNN model making an impressive £42.53 over 100 £1 bets. 
The lineup stats model's best performing ML implementation in other metrics had the worst performance with a net loss of £30.9. 
The strong performance of two ML implementations certainly gives credence to the theory that some ML models might be able to outperform the betting market. However, as can be seen the team stats model either equalled or outperformed the Lineup and stats model across all the implementations. The performance of the lineup stats model leaves room for hope however if one were to pursue with real investments the potential returns would have to be demonstrably improved with a larger dataset. 



\section{Significance and Impact}

\begin{table*}
    \centering
    \begin{tabular}{l|cccccc|c} 
         \toprule
         Model & Home & Away & Betting & Standings &  Top-4 & Relegation & Rank Sum\\ 
         \midrule
         Team Stats & 2 & 1 & 1 & 2 & 1 & 1 & 8 \\
         Lineup Stats& 1 & 2 & 2 & 4 & 1 & 1 & 11\\ 
         Players & 3 & 3 & 5 & 1 & 1 & 3 & 16\\
         Tradition & 4 & 5 & 4 & 6 & 1 & 4 & 24\\ 
         Recency & 6 & 6 & 6 & 3 & 1 & 5 & 27\\ 
         Home win & 5 & 4 & 3 & 5 & 6 & 5 & 28\\
         \bottomrule
    \end{tabular}
    \caption{Overview of model performances.}
    \label{table:matrix}
\end{table*}

    This paper explores the impact of taking lineups into account when making score predictions. The impacts of such a factor has been subjected to little interrogation in the wider literature. This fact presented as puzzling at the outset of this project, one would suspect that whether certain players play in a game has a large impact on the result. To give an example in the 2021/22 Mohamed Salah scored 23 goals.
    Liverpool as a squad scored 94 goals in the same season. When a single player can score over 24\% of a teams goals whether such players are playing in a given game should in theory have an impact on the accuracy of goals scored models. It was this intuition that this paper strived to investigate. 
    
    We performed an extensive evaluation, assessing not only the usual fitness of the models, but exploiting several real world scenarios. Table~\ref{table:matrix} shows a summary of each model's rank performance across each evaluation. 
    Overall, the machine learning methods beat the naive heuristics, and the team stats and lineup stats models together won in all categories. 
    Our results are comparable to the best ones previous found in the literature~\cite{2018INDIVIDUALTechniques}.
    Although their dataset is incompatible with our models and we could not use their model on our data, the accuracies were similar.
    
    Using lineups did not impact significantly the prediction results. However, it allowed us to explore the features analysis.
    Surprisingly, attacking stats seems not to be the most important while modelling goals scored.
    Goalkeepers clean sheets were by far and away the most impactful stat on predictive performance. 
    The justification for this seems to lie in how emblematic a goalkeepers clean sheets are for a teams defensive performance. Their clean sheets are representative of both their personal performance and the defensive performance of the team as a whole. Furthermore, it was very interesting how de-emphasised goals scored were when compared to goal creating actions. One possible justification for this might be that, once again a teams goal creating actions are representative of something larger, in this case a teams creativity. If a team is generating a lot of goal creating actions not only are they scoring a lot of goals but they are in control of how those goals are being generated. This affords the models an easier time in predicting goals scored as there is a direct link between the goals and the team's creation of them.  
    
    The final aspect of this paper's findings lie with the contextualised analysis, both the team stats and lineup stats performed exceedingly well in predicting the top 4 in the Champions League qualification zone and the bottom 3 teams relegated. 
    This seems to show that whilst accurate score prediction is still difficult to achieve outcome prediction is much more achievable. Furthermore, the betting analysis was encouraging, making any money over a period of 100 games would be difficult for the typical punter. Given that, the fact that both decision tree regression and k-nearest neighbour made significant returns of between 20\%-40\% is extremely positive and warrants further investigation. 

\balance
\bibliographystyle{acm}
\bibliography{bibliography}

\begin{thebibliography}{10}

\bibitem{Sklearn.feature_selection.chi2Documentation}
Scikit-learn 1.1.2 documentation. feature selection: Chi2.

\bibitem{WhatOffers}
{What Are The Most Common Scores In Football? | Betting Offers}.

\bibitem{arntzen2021predicting}
{\sc Arntzen, H., and Hvattum, L.~M.}
\newblock Predicting match outcomes in association football using team ratings
  and player ratings.
\newblock {\em Statistical Modelling 21}, 5 (2021), 449--470.

\bibitem{RandomDepth}
{\sc Beheshti, N.}
\newblock {Random Forest Regression}.

\bibitem{ToddMore}
{\sc Calcutt, R.}
\newblock {Todd Boehly to put data 'at the heart' of Chelsea should he
  successfully complete takeover - Sports Illustrated Chelsea FC News, Analysis
  and More}.

\bibitem{cintia2015network}
{\sc Cintia, P., Rinzivillo, S., and Pappalardo, L.}
\newblock A network-based approach to evaluate the performance of football
  teams.
\newblock In {\em Machine learning and data mining for sports analytics
  workshop, Porto, Portugal\/} (2015).

\bibitem{TheUK}
{\sc Doyle, M.}
\newblock {The 100 most expensive football transfers of all time | Goal.com
  UK}.

\bibitem{giulianotti1999football}
{\sc Giulianotti, R.}
\newblock {\em Football}.
\newblock Wiley Online Library, 1999.

\bibitem{SoccersBloomberg}
{\sc Hellier, D.}
\newblock {Soccer's Richest Game Won by Brentford and Big Data - Bloomberg}.

\bibitem{2018INDIVIDUALTechniques}
{\sc Herbinet, C.}
\newblock Predicting football results using machine learning techniques.
\newblock {\em MEng thesis, Imperial College London\/} (2018).

\bibitem{Horvat2020TheReview}
{\sc Horvat, T., and Job, J.}
\newblock {The use of machine learning in sport outcome prediction: A review}.
\newblock {\em WIREs Data Mining and Knowledge Discovery 10}, 5 (9 2020),
  e1380.

\bibitem{PredictingXplore}
{\sc Hucaljuk, J., and Rakipović, A.}
\newblock Predicting football scores using machine learning techniques.
\newblock In {\em 2011 Proceedings of the 34th International Convention
  MIPRO\/} (2011), pp.~1623--1627.

\bibitem{DecisionDepth}
{\sc K, G.~M.}
\newblock {Decision Tree Regressor explained in depth}.

\bibitem{lindberg2020comparison}
{\sc Lindberg, A., and S{\"o}derberg, D.}
\newblock Comparison of machine learning approaches applied to predicting
  football players performance.
\newblock {\em Chalmers tekniska högskola, Institutionen för data och
  informationsteknik\/} (2020).

\bibitem{Maher1982ModellingScores}
{\sc Maher, M.~J.}
\newblock {Modelling association football scores}.
\newblock {\em Statistica Neerlandica 36}, 3 (9 1982), 109--118.

\bibitem{maozad2022comparative}
{\sc Maozad, S.~N., Razali, S. N. A.~M., Mustapha, A., Nanthaamornphong, A.,
  Wahab, M. H.~A., and Razali, N.}
\newblock Comparative analysis for predicting football match outcomes based on
  poisson models.
\newblock In {\em 2022 19th International Conference on Electrical
  Engineering/Electronics, Computer, Telecommunications and Information
  Technology (ECTI-CON)\/} (2022), IEEE, pp.~1--4.

\bibitem{pacheco2015football}
{\sc Pacheco, D., de~Lima~Neto, F.~B., Moyano, L., and Menezes, R.}
\newblock Football conversations: what twitter reveals about the 2014 world
  cup.
\newblock In {\em Anais do IV Brazilian Workshop on Social Network Analysis and
  Mining\/} (2015), SBC.

\bibitem{pacheco2017using}
{\sc Pacheco, D.~F., Oliveira, M., and Menezes, R.}
\newblock Using social media to assess neighborhood social disorganization: A
  case study in the united kingdom.
\newblock In {\em The Thirtieth International Flairs Conference\/} (2017).

\bibitem{pacheco2016characterization}
{\sc Pacheco, D.~F., Pinheiro, D., De~Lima-Neto, F.~B., Ribeiro, E., and
  Menezes, R.}
\newblock Characterization of football supporters from twitter conversations.
\newblock In {\em 2016 IEEE/WIC/ACM International Conference on Web
  Intelligence (WI)\/} (2016), IEEE, pp.~169--176.

\bibitem{rajesh2020data}
{\sc Rajesh, P., Alam, M., Tahernezhadi, M., et~al.}
\newblock A data science approach to football team player selection.
\newblock In {\em 2020 IEEE International Conference on Electro Information
  Technology (EIT)\/} (2020), IEEE, pp.~175--183.

\bibitem{PremierNews}
{\sc Rojas, J.-P.~F.}
\newblock {Premier League extends {\pounds}5.1bn TV broadcast rights deal to
  2025 | Business News | Sky News}.

\bibitem{K-NearestPython}
{\sc Singh, A.}
\newblock {K-Nearest Neighbors Algorithm | KNN Regression Python}.

\bibitem{thakkar2021assessment}
{\sc Thakkar, P., and Shah, M.}
\newblock An assessment of football through the lens of data science.
\newblock {\em Annals of Data Science 8}, 4 (2021), 823--836.

\bibitem{Wu3Science}
{\sc Wu, S.}
\newblock {3 Best metrics to evaluate Regression Model? | by Songhao Wu |
  Towards Data Science}.

\end{thebibliography}

\end{document}